\pdfoutput=1
\documentclass[11pt]{article}

\usepackage{ACL2023}
\usepackage{booktabs}
\usepackage{times}
\usepackage{latexsym}
\usepackage{graphicx} 
\usepackage{subcaption}
\usepackage[T1]{fontenc}
\usepackage[utf8]{inputenc}

\usepackage{microtype}
\usepackage{tabularx} 
\usepackage{inconsolata}

\usepackage{hyperref}

\title{Query-OPT: Optimizing Inference of Large Language Models via Multi-Query Instructions in Meeting Summarization}

\author{Md Tahmid Rahman Laskar, Elena Khasanova, Xue-Yong Fu \\ \textbf{Cheng Chen, Shashi Bhushan TN} \\
          Dialpad Inc. \\
   Vancouver, BC, Canada \\ %, V6E 4A6 \\
  \texttt{\{tahmid.rahman,elena.khasanova,xue-yong,cchen,sbhushan\}@dialpad.com}}
\begin{document}
\maketitle

\begin{abstract}

%Large Language Models (LLMs) have demonstrated impressive instruction-following capabilities to solve a wide range of tasks. However, utilizing LLMs in the real world requires substantial computing resources for the majority of tasks. 

This work focuses on the task of query-based meeting summarization, in which the summary of a context (meeting transcript) is generated in response to a specific query. When using Large Language Models (LLMs) for this task, usually a new call to the LLM inference endpoint/API is triggered for each new query, even if the context stays the same. However, repeated calls to the LLM inference endpoints would significantly increase the costs of using them in production, making LLMs impractical for many real-world use cases. To address this problem, in this paper, we investigate whether combining the queries for the same input context in a single prompt to minimize repeated calls can be successfully used in meeting summarization. In this regard, we conduct extensive experiments by comparing the performance of various popular LLMs: GPT-4, Gemini-1.5, Claude-3, LLaMA-2, Mistral, Phi-3, and Qwen-2 in single-query and multi-query settings. We observe that % the capability to ensure 
100\% reliability in generating the response in the expected format is usually limited to certain closed-source LLMs, with most open-source LLMs lagging behind (except a few 7B parameters LLMs like Mistral and Phi-3).  
We conclude that multi-query prompting could be useful to significantly optimize the inference costs %by significantly reducing calls to the inference APIs 
%for the task of 
in meeting summarization.

\end{abstract}
%\vspace{-1mm}
\section{Introduction}
 %\vspace{-1mm}
One key advantage of recent LLMs is their %strong emergent abilities\footnote{The emerging behavior of language models refers to certain capabilities that are only present in larger language models but not in smaller language models \cite{wei2022emergent}} \cite{wei2022emergent}, which helps them to achieve 
impressive instruction-following capabilities even in zero-shot scenarios (without fine-tuning on a particular task) \cite{laskar2023systematicchatgpt,qin2023chatgpt,bang2023multitask}. This instruction-following ability of LLMs has also led to an increase in utilizing LLMs for many real-world use cases \cite{laskar-etal-2023-building}. %For instance, in tasks like meeting summarization, LLMs have been widely utilized in recent times due to their impressive zero-shot performance \cite{laskar-etal-2023-building}.  
However, despite their impressive performance, deploying LLMs in the real world is not trivial, with one major obstacle being significant inference costs. Thus, optimizing the inference cost while ensuring high accuracy and efficiency is important for practical applications. % settings. 

Although several optimization techniques  \cite{zhu2023survey}, such as knowledge distillation, post-training quantization, etc. are utilized to minimize the cost associated with LLMs in production, these techniques cannot be applied to the closed-source LLMs like GPT-4 \cite{openai2023gpt4} or Gemini \cite{team2023gemini}. For open-source LLMs \cite{touvron2023llama2}, these techniques may come with different trade-offs. For instance, while quantization \cite{zhu2023survey}  can reduce the GPU requirement and make it possible to do LLM inference on less expensive hardware, it may also result in slower inference speed and reduced accuracy. 

Another limitation is that the cost\footnote{\url{https://huggingface.co/spaces/philschmid/llm-pricing}} associated with LLM inference depends on the number of tokens processed by LLMs, which is true for both closed-source and open-source LLMs. This issue can be addressed by reducing either the context size or the number of calls to the inference endpoints. For the task of query-focused meeting summarization, the latter is highly preferable since there can be many queries that may require the full context of long meeting transcripts and so truncating the context size could lead to a drop in performance. %s it can address the limits on API calls by enforcing the model to process the same large input contexts only once. %, which often results in performance gain.

%To address the above issues, 
In this paper, we study how we can minimize calls to the LLM inference APIs/endpoints in the meeting summarization task by optimizing the prompts. More specifically, we investigate whether we can combine the queries for the same context in a single prompt to minimize calls to the same inference endpoints for different query-context pairs. For this purpose, we created a new version of the QMSUM dataset \cite{zhong2021qmsum} for the query-focused meeting summarization task by combining all queries for the same context in a single prompt. We conduct extensive experiments with several open-source and closed-source LLMs and compare their performance in both single-query and multi-query versions of the dataset.
Our experimental results show that only certain closed-source LLMs were able to reliably answer all the queries given in a single prompt in the required format in the multi-query setting. Meanwhile, most open-source LLMs, even after fine-tuning, fail to properly follow multi-query instructions to generate the response in the requested format. We also find a similar trend in zero-shot scenarios in some larger closed-source LLMs.  %However, one notable exception in this regard is GPT-4, which demonstrates better instruction-following capabilities to generate summaries for multiple queries from meeting transcripts. %However, a notable exception is a fine-tuned FLAN-T5-Large, which achieves performance on par with much larger LLMs (from 7B to more than 70B) used in zero-shot settings, while being significantly smaller. This makes smaller LLMs like FLAN-T5 a suitable cost-efficient LLM for real-world deployment.
Our extensive experiments demonstrate the capabilities and limitations of different LLMs in following multi-query instructions in meeting summarization. This gives strong insights into utilizing LLMs in real-world settings to minimize the inference cost for similar applications. % in building efficient real-world %query-focused meeting summarization 
%LLM-powered systems. % to handle multiple queries. %for multi-query meeting summarization. %Thus, providing valuable guidance for similar industrial applications. Considering various factors, we employ LLaMA-2-7B in a real-world setting to generate summaries from Automatic Speech Recognition (ASR)-generated transcripts \cite{fu-etal-2022-entity,khasanova2022developing,laskar2022auto,laskar-etal-2022-blink,laskar-etal-2023-ai-coach-assist,manderscheid-lee-2023-predicting} of organizational meetings. 
Our major contributions are as follows:

(1) We conduct an extensive evaluation of various LLMs in the multi-query setting for query-focused meeting summarization to investigate their capability in following multi-query instructions in comparison to the traditional single-query scenario. % of organizational meetings. ASR-generated

(2) We observe that while most LLMs demonstrate the ability to respond to multiple queries in a single prompt, many of these LLMs could not achieve 100\% reliability in generating the responses in the required format (with the exception being certain closed-source LLMs). %To ensure fair evaluation, we propose a new evaluation criteria based on ROUGE where all the query-summary pairs for the same transcript are combined to compute the score.

(3) The findings from our experimental evaluation will provide insights into optimizing prompts to reduce production costs while deploying LLMs for real-world usage. As a secondary contribution, we will release our constructed multi-query version of the QMSUM dataset and the code here: \url{https://github.com/talkiq/dialpad-ai-research/tree/main/query_opt}. % to help accelerate further research.

%his paper thus presents a detailed examination of the use of LLMs for the specific task of meeting summarization, offering a practical perspective on the trade-offs that come with selecting a model based on its performance, cost, and computational requirements. The results and insights we share in this paper serve to inform and guide researchers, practitioners, and industry professionals alike in their quest to harness the power of LLMs for efficient meeting summarization.

%\vspace{-1mm}
\section{Related Work}
%\vspace{-1mm}
The impressive instruction-following capabilities of LLMs have led to their wide adoption in the real world for various tasks, which includes generating summaries from meeting transcripts \cite{laskar-etal-2023-building}. However, in many scenarios, users may require extracting other information from the transcripts rather than a generic summary of the meetings. In such cases, one straightforward way is to call the LLM inference API/endpoint for the given query-transcript pair. However, this approach is not cost-effective, since the same transcript for a different query would be given as input again to the LLM in different calls. % to the LLM inference endpoint. 
Thus, it will lead to a non-optimal usage cost for processing the same tokens in a transcript multiple times.

One possible solution in this regard could be combining the queries in a single prompt, similar to the work of \citet{laskar2023systematicchatgpt} where they evaluated ChatGPT\footnote{\url{https://openai.com/chatgpt}} (i.e., GPT-3.5) in the open-domain question-answering task in about 100 samples from Natural Questions \cite{nqopen} and WebQuestions \cite{berant2013webquestions} datasets. While their evaluation shows that instruction-following LLMs like GPT-3.5 can respond to multiple queries in a single prompt, they did not investigate the following research questions: 

(i) \textit{Are LLMs capable of responding to multiple questions in a given input text that requires understanding of long conversation context?} 

(ii) \textit{Can LLMs generate the response in a specified format to ensure easier parsing of the output?} 

(iii) \textit{Do smaller open-source LLMs also possess the ability to respond to multiple queries in a single prompt similar to larger closed-source LLMs such as ChatGPT?}

To address the above questions, in this paper, we conduct a comprehensive evaluation of popular closed-source and open-source LLMs in the QMSUM \cite{zhong2021qmsum} dataset for query-focused meeting summarization task to investigate their performance in following multi-query instructions to extract information from long conversations.  
% \footnote{The smaller open-source LLMs were not evaluated for multi-query instructions by \cite{laskar2023systematicchatgpt}.}
%\vspace{-1mm}
\section{Our Methodology}
%\vspace{-1mm}

\begin{figure*}[t!]
    \centering

        \includegraphics[width=\linewidth]{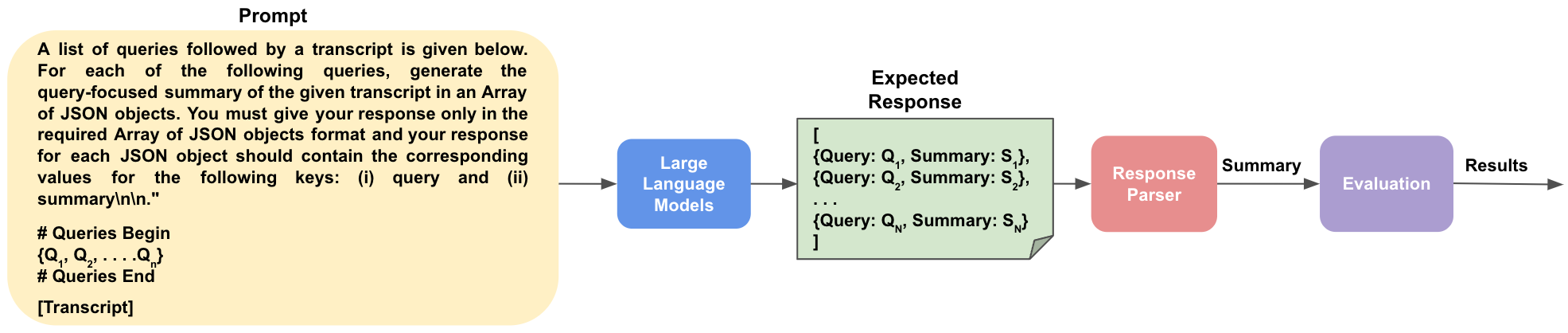}
     %   \caption{Based on different model types.}
      %  \label{fig:average_results_instruction}
    
%\vspace{-1mm}
 
    \caption{\small{An overview of our Multi-Query Instruction Evaluation Framework for the Query Focused Summarization Task.}}
    \label{fig:overview}
\end{figure*}

In this section, we present our overall methodology to evaluate the multi-query instruction capabilities of LLMs. Below, we describe our dataset construction procedure, evaluation approach, and the models used in these experiments.

%\vspace{-1mm}
\subsection{Dataset} %\vspace{-1mm}
The objective of this research is to study whether LLMs are capable of following multi-query instructions to extract information from the given source text depending on the input queries. For this purpose, we utilize the QMSUM dataset \cite{zhong2021qmsum} and convert it to a multi-query instruction dataset for query-focused meeting summarization. The original dataset consists of query-transcript pairs, with the same transcript appearing multiple times for different queries. In our modified multi-query instruction version of the QMSUM dataset, we combine all the queries for the same input transcript to construct a single prompt. More specifically, for the input transcript $T$, we combine the queries $q_1$, $q_2$, . . . $q_n$ to construct the multi-query set $Q$. Then, we merge it with the instruction $I$ that explains the task and the required output format. This results in reduced samples in the multi-query version of QMSUM: \textbf{162/35/35} instances in train/validation/test sets, whereas the original dataset has  \textbf{1257/272/281} instances, respectively. 

%\vspace{-1mm}
\subsection{Evaluation Framework} %\vspace{-1mm} 
For each sample in the multi-query instruction format in the dataset, at first, the response is generated by the respective LLM for the given multi-query input. Then we parse the output to extract the summary for each corresponding query from the query-summary pairs from the generated response. %If a summary cannot be parsed for a given query (i.e., the output is not generated in the required format), we consider that the LLM fails to generate a summary for that specific query. In this case, we consider it as a ``NONE'' response. 
Finally, we evaluate the model's performance according to several criteria described in Section \ref{sec:results}. % across the whole dataset. 
An overview of our proposed multi-query instruction framework is shown in Figure \ref{fig:overview}.

%\vspace{-1mm}

\subsection{Models} 
%\vspace{-1mm}
%In this section, we describe the models that we study in this work. 
Since the QMSUM dataset has on average about 9K words per transcript  \cite{zhong2021qmsum}, which is approximately 12K tokens\footnote{100 tokens are equivalent to 75 words: \url{https://platform.openai.com/tokenizer}}, only the LLMs that can handle longer contexts (e.g., support at least 20K tokens) are selected. We set the \textit{maximum output tokens} limit to 2000 to allow enough token count for the multi-query responses and also set the \textit{maximum input tokens} limit to 20000 to effectively utilize long conversation context. %for the open-source models to meet their input size limits. 
Apart from setting the input/output token length, we use the temperature value of 1.0 and default values for other decoding parameters from HuggingFace \cite{wolf2019huggingface} for open-source LLMs and respective API providers for closed-source LLMs. 
Since this research aims to ensure efficiency in real-world LLM inference, we do not select any open-source LLMs with more than 7B parameters. Below, the models that are studied in this paper are described.   %Open-source LLMs are implemented using HuggingFace. % \cite{wolf2019huggingface}. 

\textbf{GPT-4:} It is the most powerful LLM released by OpenAI that also currently powers ChatGPT and achieves the best performance in several benchmarks   \cite{openai2023gpt4}. %\footnote{\url{https://lmsys.org/blog/2023-06-22-leaderboard/}} 
 %In several benchmarks, GPT-4 is found to be the best-performing LLM. % ChatGPT and obtains impressive zero-shot performance across various tasks \cite{laskar2023systematicchatgpt}. %There are two versions of GPT-3.5 being released, with context windows of 4K and 16K tokens. However, the 16K tokens version is 2 times more expensive than the 4K tokens version. Considering the cost, 
We use the \textit{gpt-4o} %\textit{gpt-4-0125-preview}, 
and the \textit{gpt-4-turbo} models\footnote{\url{https://platform.openai.com/docs/models}} in this work. % with the default parameters from OpenAI\footnote{\url{https://platform.openai.com/docs/models/}}. 

%\textbf{GPT-3.5:} %It is an autoregressive LLM that leverages reinforcement learning from human feedback (RLHF) mechanism. %Contrary to the earlier models released by OpenAI, it has been additionally trained using dialog-based instructional data, leading to better 
%It is the first backbone model behind ChatGPT and obtains impressive zero-shot performance across various tasks \cite{laskar2023systematicchatgpt}. %There are two versions of GPT-3.5 being released, with context windows of 4K and 16K tokens. However, the 16K tokens version is 2 times more expensive than the 4K tokens version. Considering the cost, 
%We use the \textit{gpt-3.5-turbo-0125} model. % with the default parameters from OpenAI. 

\textbf{Gemini-1.5:} LLMs in the Gemini \cite{team2023gemini} family are developed by Google and is currently considered the most advanced LLM in comparison to other LLMs \cite{palm2} offered by Google. We use the \textit{gemini-1.5-pro} model that ensures advanced reasoning capability and the \textit{gemini-1.5-flash} model optimized for inference efficiency. %ensures inference speed efficiency with reduced usage cost. %\textit{Google's VertexAI\footnote{\url{https://cloud.google.com/vertex-ai/docs/generative-ai/model-reference/text}}} % with the default parameters for PaLM-2. 

{\textbf{Claude-3:} The Claude-3 family\footnote{\url{https://www.anthropic.com/claude-3-model-card}}  (Haiku, Sonnet, and Opus) LLMs are introduced by Anthropic. %which demonstrated impressive performance in processing speed and reasoning capabilities. 
We use the \textit{Claude-3-Haiku} model which is cost and speed-optimized, the \textit{Claude-3-Opus} model which achieves the best result in terms of reasoning capability, and the recently proposed \textit{Claude-3.5-Sonnet}\footnote{\url{https://www.anthropic.com/news/claude-3-5-sonnet}} model.

\textbf{LLaMA-2:} LLaMA-2 \cite{touvron2023llama2} is an open-source LLM developed by Meta which is one of the pioneer open-source LLMs available. % for both commercial and research purposes.
We could not use the most advanced version in the LLaMA series, the LLaMA-3  \cite{dubey2024llama3herdmodels} model, since it does not support more than 8K tokens. While the LLaMA-2 model is also limited to 4K tokens, we use its long context variant, the \textit{LLaMA-2-7B-32K-Instruct}\footnote{\url{https://huggingface.co/togethercomputer/LLaMA-2-7B-32K}} model from Together. 

\textbf{Mistral:} The Mistral series models \cite{jiang2023mistral,jiang2024mistral8x7b} are proposed by Mistral AI. It leverages grouped-query and sliding window attention to effectively handle long sequences. % of arbitrary length. 
We use its instruction-tuned \textit{Mistral-7B-instruct-v0.3} model. % and the Mixtral 8x7B model that leverages the sparse Mixture of Experts technique \cite{cai2024survey}.

%This leads to an improved performance. % This architectural change has led it to outperform LLaMA-2-70B in several benchmarks. %for each token to have access to 47B parameters while using only 13B active parameters during inference. We use it for zero-shot evaluation with its default parameters. 

\textbf{Qwen2:} The Qwen2 series LLMs \cite{bai2023qwen,yang2024qwen2technicalreport} from Alibaba support long context lengths and achieve strong performance on various %Chinese and English 
benchmarks, outperforming other open-source LLMs. % like LLaMA-3 and Mistral. 
We use the \textit{Qwen2-7B-Instruct}\footnote{\url{https://huggingface.co/Qwen/Qwen2-7B-Instruct}} model in this paper.

\textbf{Phi-3:} Microsoft's Phi-3 series LLMs \cite{abdin2024phi} include models from 3 billion to 14 billion parameters. We use the \textit{Phi-3-mini-128k-instruct} (3B) model and the \textit{Phi-3-small-128k-instruct} (7B) from HuggingFace\footnote{\url{https://huggingface.co/docs/transformers/main/en/model_doc/phi3}} in this work.

%\textbf{FLAN-T5-XL:} FLAN-T5 \cite{flant5} is an extension of the T5 \cite{t5} model. %The T5 model treats each task as a sequence-to-sequence problem. 
%While the architecture of FLAN-T5 is similar to the original T5 model, it leverages instruction fine-tuning instead of traditional fine-tuning. %We use . % in our experiments with the learning rate set to $2e-5$. We run 10 epochs for FLAN-T5-Large and 20 epochs for Base and Small. %(small and base) and 

%\vspace{-1mm}
\section{Experimental Results}
\label{sec:results}
%\vspace{-1mm}
%In this section, we first present the LLMs used for evaluation, %then followed by the performance metrics achieved by the models. 

%\subsection{Results and Discussions}
%\vspace{-1mm}
In this section, we present our experimental findings in the multi-query setting to investigate the following: (i) {LLMs capability to generate responses in the expected format}, (ii) {effectiveness when compared with single-query settings}, (iii) {effects of fine-tuning}, (iv) 
{qualitative evaluation of the generated summaries}, (v) performance robustness, and (vi) usage cost analysis. 

\begin{figure}[t!]
    \centering

        \includegraphics[width=\linewidth]{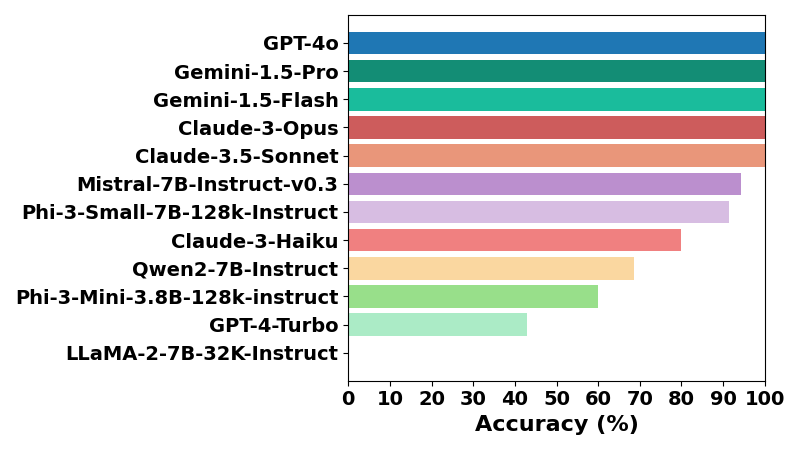}
     %   \caption{Based on different model types.}
      %  \label{fig:average_results_instruction}
    
%\vspace{-3mm}
 
    \caption{\small{Format following capability of LLMs in zero-shot.}}
    \label{fig:format_accuracy}
\end{figure}

%\vspace{-1mm}
\subsection{Format Following Capability of LLMs}
%\vspace{-1mm}

At first, we investigate whether LLMs could properly generate the output in the required JSON format. This is important since we need to extract the summary for each query given in the multi-query input. Therefore, we report the accuracy in terms of accurately generating the response in the expected JSON format in Figure \ref{fig:format_accuracy} and surprisingly find that many LLMs could not generate the response (see Appendix \ref{appendix} for sample responses) in the expected JSON format or in a consistent format. Here, consistency in formatting refers to such scenarios when a response is not in the correct JSON format, but the summaries for the corresponding queries can be parsed easily using the parser by applying some post-processing techniques that can be generalized across all responses, i.e.,  do not cause any issues while parsing other responses. Our evaluations demonstrate that while most of the closed-source LLMs have 100\% accuracy in generating the response in the required format (except Claude-3-Haiku with about 80\% accuracy and GPT-4-Turbo having less than 50\% accuracy), none of the open-source LLMs could achieve 100\% reliability in format following. Among open-source LLMs, Mistral-7B-Instruct-V0.3 is found to be the best, achieving around 95\% accuracy while outperforming larger closed LLMs like Claude-3-Haiku and GPT-4-Turbo. These closed-source LLMs are also outperformed by the Phi-3-Small model which achieves around 90\% accuracy. Meanwhile, both the Qwen2-7B-Instruct and the Phi-3-Mini models fail to obtain more than 70\% accuracy. We also surprisingly find that GPT-4-Turbo makes errors in generating the response in the expected format in more than 50\% of the cases, with the LLaMA-2-7B-32K being fully unable to generate the response in the required format, having 100\% error.  

%\vspace{-1mm}
\subsection{Performance of LLMs in Multi-Query and Single-Query Settings in Zero-Shot}
\label{sec:performance}
%\vspace{-1mm}
For performance evaluation in multi-query and single-query settings, we follow prior work in query-focused meeting summarization and report the results based on the commonly used evaluation metrics, namely, ROUGE-1, 2, L scores \cite{lin2004rouge}, and the BERTScore \cite{zhang2019bertscore} based on the \textit{DeBERTa-xlarge-mnli }\cite{he2020deberta} model. In addition, we use the AlignScore \cite{zha-etal-2023-alignscore} metric\footnote{\url{https://huggingface.co/yzha/AlignScore/resolve/main/AlignScore-large.ckpt}} to evaluate the factual consistency of the LLM-generated summaries. %In addition, we use the AlignScore metric \cite{zha-etal-2023-alignscore} to evaluate the factual consistency of the model-generated summaries. 

Intuitively, we would only be interested in evaluating the summary responses and will thus need to extract the corresponding summary for each query from the generated response. However, it is challenging to do so since many LLMs are not 100\% reliable in generating the response in the required format. In many cases, even after applying advanced post-processing, % on LLM-generated responses, 
it was not possible to extract the required summary for the corresponding query. %If the parsing script cannot extract the required summary, it won't be possible to provide the summary for the respective query in real-world inference environments. 
In such cases, we consider the summaries that cannot be parsed for respective queries as empty responses.  We present the results for both single-query and multi-query settings in Table \ref{tab:results_all}. Below, we summarize our observations: %while evaluating 
%in the multi-query setting. %Therefore, we propose the following evaluation procedure based on ROUGE scores: 

\begin{table*}[t!]
\centering
\scriptsize
\setlength{\tabcolsep}{1pt}
\begin{tabular}{l|ccccc|ccccc}
\toprule
%\multicolumn{1}{c}{} & \multicolumn{6}{c}{\textbf{Dataset}} \\ 
%\cmidrule(r){2-7}
%\multicolumn{1}{c}{} & \multicolumn{8}{c}{\textbf{QMSUM}} %& \multicolumn{4}{c}{\textbf{QMSUM-I}} 
%\\
%\cmidrule(r){2-9}
\multicolumn{1}{c}{} & \multicolumn{5}{c}{\textbf{Multi-Query}} %& \multicolumn{4}{c}{\textbf{Single-Query}} & \multicolumn{4}{c}{\textbf{Multi-Query}} 
& 
\multicolumn{5}{c}{\textbf{Single-Query}}  \\ 
\cmidrule(r){1-1}
\cmidrule(r){2-6} \cmidrule(r){7-11} % \cmidrule(r){10-13} \cmidrule(r){14-17}
%\multicolumn{1}{c}{\textbf{Models}}
%& \multicolumn{3}{c}{\textbf{ROUGE}} & \multicolumn{1}{c}{\textbf{BERTScore}} & \multicolumn{3}{c}{\textbf{ROUGE}} & \textbf{BERTScore} \\ %& \textbf{R-1} & \textbf{R-2} & \textbf{R-L} & \textbf{B-S} & \textbf{R-1} & \textbf{R-2} & \textbf{R-L} & \textbf{B-S}  \\
%\cmidrule(r){2-4} \cmidrule(r){6-8}
%& \textbf{1} & \textbf{2} & \textbf{L} & \multicolumn{1}{c}%{\textbf{}}  & \textbf{1} & \textbf{2} & \textbf{L} & \textbf{} \\ 

{\textbf{Models}} & \multicolumn{1}{c}{\textbf{ROUGE-1}}  & \multicolumn{1}{c}{\textbf{ROUGE-2}}  & \multicolumn{1}{c}{\textbf{ROUGE-L}}  & {\textbf{BERTScore}}  & {\textbf{AlignScore}} & \multicolumn{1}{c}{\textbf{ROUGE-1}}  & \multicolumn{1}{c}{\textbf{ROUGE-2}}  & \multicolumn{1}{c}{\textbf{ROUGE-L}} & \multicolumn{1}{c}{\textbf{BERTScore}} & {\textbf{AlignScore}} \\
\midrule
\textbf{GPT-4o}& 31.6 & 8.1 & 20.1 & 62.8 & 24 & 26.8 & 7.2 & 16.4 & 58.1 & 15 \\

%N/A & N/A & N/A & N/A & N/A & N/A & N/A & N/A \\
%34.4 & 11.6 & 20.1 & 64.5 & 46.3 & 14.8 & 23.4 & 66.4 \\

%\textbf{GPT-4-0125-Preview} & 30.9 & 7.3 & 19.2 & 61.7 & & 24.1 & 6.0 & 15.0 & 56.9 \\  

%N/A & N/A & N/A & N/A & N/A & N/A & N/A & N/A \\
%24.1 & 6.0 & 15.0 & 56.9  & 24.1 & 6.0 & 15.0 & 56.9 \\

\textbf{GPT-4-Turbo} & 11.1 & 2.5 & 7.0 & 41.0 & 12 & 24.6 & 5.7 & 15.0 & 57.2 & 15 \\

%N/A & N/A & N/A & N/A & N/A & N/A & N/A & N/A  \\
%& 24.6 & 5.7 & 15.0 & 57.2 & 24.6 & 5.7 & 15.0 & 57.2  \\

\textbf{Gemini-1.5-Pro}& 31.2 & 8.0 & 19.7 & 61.5 & 21 & 29.3 & 7.5 & 18.0 & 59.4 & 13 \\

%& 28.2 & 8.3 & 16.4 & 61.8 & 43.2 & 12.5 & 21.1 & 63.9 \\

\textbf{Gemini-1.5-Flash} & \textbf{33.4} & \textbf{9.5} & \textbf{21.6} & \textbf{62.9} & 22 & \textbf{30.9} & \textbf{8.6}  & \textbf{19.6} & \textbf{60.5} & 14 \\

%& 30.9 & 8.6  & 19.6 & 60.5 & 30.9 & 8.6  & 19.6 & 60.5 \\

\textbf{Claude-3-Opus} & 33.3 & 9.4 & 21.2 & 62.5 & 18 & 25.4 & 7.2  & 15.6 & 55.3 & 15 \\

%& 25.4 & 7.2  & 15.6 & 55.3 & 25.4 & 7.2  & 15.6 & 55.3 \\

\textbf{Claude-3.5-Sonnet} & 32.3 & 8.6 & 20.2 & 62.0 & 21 & 25.0 & 7.0  & 15.3 & 54.4 & 17 \\

%& 25.0 & 7.0  & 15.3 & 54.4 & 25.0 & 7.0  & 15.3 & 54.4 \\

\textbf{Claude-3-Haiku} & 26.0 & 7.1 & 16.7 & 55.2 & 15 & 25.4 & 6.9  & 15.7 & 55.7 & 16 \\

%& 25.4 & 6.9  & 15.7 & 55.7 & 25.4 & 6.9  & 15.7 & 55.7 \\

\midrule
\textbf{Mistral-7B-Instruct-v0.3} & \textbf{30.1} & \textbf{9.4} & \textbf{20.3} & \textbf{59.8} & 18 & \textbf{26.4} & \textbf{7.2} & \textbf{17.0} & \textbf{58.2} & 14 \\

%& 26.4 & 7.2 & 17.0 & 57.2 & 26.4 & 7.2 & 17.0 & 57.2 \\

%\textbf{Mixtral-8x7B-Instruct-v0.1}& 26.1 & 7.4 & 17.1 & 56.2 & 26.2 & 6.9 & 16.9 & 58.1 \\

%& 26.2 & 6.9 & 16.9 & 58.1 & 26.2 & 6.9 & 16.9 & 58.1 \\

\textbf{Qwen2-7B-Instruct}& 17.4 & 4.3 & 11.1 & 47.6 & 15 & 9.7 & 2.3 & 6.1 &  44.2 & 7 \\

%& 9.7 & 2.3 & 6.1 &  44.2 & 9.7 & 2.3 & 6.1 &  44.2 \\

\textbf{Phi-3-Small-7B-128K-Instruct}& 28.8 & 7.7 & 18.8 & 59.5  & 11 & 23.7 & 5.8  & 15.2 & 56.9 & 9 \\ 

%23.7 & 5.8  & 15.2 & 56.9 & 23.7 & 5.8  & 15.2 & 56.9 \\
\textbf{Phi-3-Mini-3.8B-128K-Instruct}& 18.4 & 4.7 & 11.8 & 47.9 & 11 & 22.5 & 5.3 & 14.1 & 56.0 & 8 \\ 

%22.5 & 5.3 & 14.1 & 56.0 & 22.5 & 5.3 & 14.1 & 56.0 \\
\textbf{LLaMA-2-7B-32k-instruct}& 0.0 & 0.0 & 0.0 & 0.0 & 0 & 10.3 & 2.2 & 6.8 & 40.7 & 5 \\

%& 0 & 0 & 0 & 0 & 2.2 & 6.0 & 13.4 & 49.4  \\
% N/A &
\bottomrule
% %\vspace{-3mm}
\end{tabular}
%\vspace{-1mm}
\caption{\small{Performance of LLMs on the QMSUM dataset with multi-query and single-query prompting in zero-shot settings.}}
\label{tab:results_all}
\end{table*}

% \textit{- For each transcript, instead of parsing the query and the summary from the generated response, we compute the ROUGE score between multiple summaries generated for a transcript for the given multi-query input with the combined reference summaries of the given transcript. }
%\textit{- In the multi-query scenario, for each transcript, we compute the ROUGE score between the reference, which combines summaries for each query (with the queries removed), and the entire output returned by the model.}

%\textit{- To ensure a fair comparison between the models in multi-query and single-query settings, we similarly combine all output summaries for a given transcript in the single-query scenario and compute ROUGE scores against the combined reference summaries. }

 %To investigate the effect of multi-query prompting, we compare the results with the traditional single-query prompting setting. 

    (i) In the zero-shot setting, we find that in general, most closed-source LLMs outperform the open-source ones in both multi-query %scenario. % across all metrics. 
    and single-query scenarios. %, their performance is usually more comparable. % in most cases. 
    
    (ii) Overall, we find that Gemini-1.5-Flash performs the best across all metrics in both single and multi-query scenarios, outperforming all other closed-source and open-source LLMs. Nonetheless, other closed-source LLMs like Claude-3-Opus, Claude-3.5-Sonnet, Gemini-1.5-Pro, and GPT-4o also achieve comparable performance in multi-query settings. In terms of open-source LLMs, Mistral-7B-Instruct-v0.3 achieves the best results in both single and multi-query scenarios. 

    (iii) We find that LLMs that demonstrate higher accuracy in the format following perform poorer in single-query settings than in multi-query settings in the zero-shot experiment. Our investigation shows that LLMs used in single-query scenarios tend to generate longer summaries than gold summaries, resulting in poorer performance in terms of ROUGE and BERTScore. However, in multi-query settings, due to the maximum output token limit, the average length of the summary is closer to the reference, which is reflected in better scores. For instance, for each query, while the average gold summary length is 64.7 words, the average summary length of LLMs that showed 100\% format following accuracy is 73.9 words in multi-query settings and 162.4 words in single-query settings. 

    (iv) To investigate whether the performance difference in single-query and multi-query settings for different models is statistically significant or not, we conduct paired t-test (p $\le$ 0.05) and find that they are not statistically significant. This is possibly due to large discrepancies in the format following performance between the models in the multi-query scenario. However, when we only consider the models that achieve at least 80\% format following accuracy in multi-query settings, we find that the performance difference is statistically significant in terms of ROUGE-1 and BERTScore.
    
    %generate longer summaries, which result in poorer ROUGE scores in single-query scenarios, 

  %  (iii) Interestingly, we find that the performance of each fine-tuned model in single-query settings is much better than zero-shot LLMs. This could be due to the fact that fine-tuning helps LLMs to better understand the required summary length, resulting in an improved ROUGE score.
%\vspace{-1mm}

\begin{table}[t!]
\scriptsize
\setlength{\tabcolsep}{2.5pt}
    \centering
    \begin{tabular}{l|cccc}
        \toprule
        \textbf{Model Name} & \textbf{ROUGE-1} & \textbf{ROUGE-2} & \textbf{ROUGE-L} & \textbf{BERTScore} \\ % & %\textbf{FFC}  \\
        \midrule
      %   LLaMA-2-7B-32K & 30.4 ( 30.4) & 2.9 ( 2.9) & 1.1 ( 1.1) & 2.1 ( 2.1) \\
     %   Qwen2-7b-instruct & 34.3 ( -13.3) & 5.6 ( -11.8) & 1.7 ( -2.6) & 3.8 ( -7.3) \\
        \textbf{Mistral-7B-Instruct}& 30.0 (-0.1) & 10.1 (+0.7) & 20.9 (+0.6) & 59.9 (+0.1)  \\

    %   \textbf{Phi-3-Small-Instruct} & 24.9 (+2.9) & 8.3 (+2.2) & 18.3 (+3.9) & 53.2 (+1.2)  \\

        \textbf{Phi-3-Mini-Instruct} & 20.1 (+1.7) & 6.7 (+2.0) & 14.5 (+2.7) & 48.4 (+0.5)  \\
        \bottomrule
        
    \end{tabular}
    \caption{\small{Results for some models after fine-tuning in multi-query scenarios. The loss and gain in performance compared to the zero-shot results for ROUGE and BERTScore mentioned in Table \ref{tab:results_all} are demonstrated inside brackets.}}
    \label{tab:results_fine_tuning}
\end{table}

%\vspace{-1mm}
\subsection{Effects of Fine-Tuning}

%\vspace{-1mm}
We have found in our prior experiments that open-source LLMs in zero-shot scenarios usually fail to achieve 100\% format following accuracy. In this section, we investigate whether fine-tuning the open-source models could improve the performance of the following models: Mistral-7B-Instruct-v0.3 and Phi-3-Mini-128K-Instruct. We conduct full fine-tuning of these models with the learning rate being set to $2e-5$, batch size = 1, epochs = 10, and max input/outputs tokens = 20000/2000. We selected the model for evaluation on the test set that performs the best on the validation set in a particular epoch. In terms of format-following capability, while for Mistral-7B, the accuracy remains the same (accurate in about 95\% of the cases), we observe a 9\% gain for Phi-3-Mini. We show the summarization performance for these 2 LLMs in Table \ref{tab:results_fine_tuning} to find that the overall ROUGE scores and the BERTScore are generally increased for both LLMs. %While the gain is not that much for Mistral which is already very good in zero-shot, for the Phi-3 models the improvement is by a higher margin (e.g., fine-tuning improving the ROUGE-1 score by 2.9\% for Phi-3-Small and by 1.7\% for Phi-3-Mini). 

%  ($\Delta$performance gain from zero-shot)

% \begin{table}[t!]
%\centering
%\tiny
%\setlength{\tabcolsep}{2pt}
%\begin{tabular}{c|c|c|c}
%\toprule %0.7009676372					0.6792449793						0.7009676372	0.6899353658
  %  \textbf{Model} & 
% \textbf{Type} & \textbf{ROUGE-1} & \textbf{ROUGE-2} & \textbf{ROUGE-L}\\ 
%\midrule
%GPT-3.5 ($n$ = 5000) &         truncation (2 Sentence) & 31.48 & 13.32 & 23.64 & 60.25 &         49.60\\
 %\textbf{Multi-Query} & 30.9 & 7.3 & 19.2 \\ 
 %\textbf{Single-Query} & 24.1 & 6.0 & 15.0 \\

%\bottomrule
%\end{tabular}

%\vspace{-1mm}
%\caption{\small{Case study results after parsing GPT-4 generated responses \textit{{without}} combining reference summaries.}}
%\label{table:case_study}
%\end{table}

\subsection{Human Evaluation}

%\vspace{-1mm}
In this section, we present our findings by conducting human evaluation in two settings: (i) Qualitative Evaluation, and (ii) Preference Test.

\paragraph{Qualitative Evaluation:} %We conducted the qualitative evaluation of the LLM-generated summaries in our proposed multi-query settings. 
For this purpose, we follow the prior work in query-focused text summarization to conduct a qualitative evaluation \cite{laskar-etal-2022-domain} on the LLM-generated responses in the multi-query setting across randomly sampled 10 conversations\footnote{Only those samples were selected where LLMs could accurately generate the response in the required format} for each of the corresponding queries. We evaluate Fluency, Coherence, Informativeness, and Factual Correctness. All the samples were annotated by 2 human annotators having expertise in computational linguistics. The human annotators' ratings are averaged and presented in Table \ref{tab:results_human_eval} for the following LLMs\footnote{We select those LLMs that achieve more than 90\% format following accuracy and sufficient ROUGE and BERTScore.}: GPT-4o, Gemini-1.5 (Pro and Flash), Claude-3-Opus, Claude-3.5-Sonnet, Mistral-7B-Instruct-V3 (both zero-shot and fine-tuned). Based on the results, it is evident that none of the LLMs struggled with Fluency. Further, all closed-source LLMs usually maintain high Coherence, with Gemini-1.5-Pro achieving the best performance in this metric. However, the performance of open-source LLMs on Coherence is notably below the closed-source ones, which is also observed in terms of Informativeness and Factuality. While all closed-source LLMs achieve higher Factual Correctness scores, the Informativeness score for all closed-source LLMs is comparatively lower, which we also observe for open-source LLMs. More specifically, the open-sourced Mistral models achieve quite poor Informativeness scores (e.g., below 3.0). In general, similar to the automatic evaluation, closed-source LLMs again achieve better results. %demonstrate better performance than open-source ones based on human evaluation. %, which we also observe in terms of automatic metrics. 

\begin{table}[t!]
\scriptsize
\setlength{\tabcolsep}{1.5pt}
    \centering
    \begin{tabular}{l|cccc}
        \toprule
        \textbf{Model Name} & \textbf{Fluency} & \textbf{Coherence} & \textbf{Informativeness} & \textbf{Factuality} \\ % & %\textbf{FFC}  \\
        \midrule
        \textbf{GPT-4o}& 4.9 & 4.4 & 4.0 & 4.7  \\
    
       \textbf{Gemini-1.5-{Pro}} & 4.9 & 4.9 & 4.3 & 4.8  \\

        \textbf{Gemini-1.5-Flash} & 4.6 & 4.5 & 4.2 & 4.7  \\

\textbf{Claude-3-Opus} & 4.7 & 4.6 & 4.1 & 4.8  \\

        \textbf{Claude-3-Sonnet} & 4.7 & 4.6 & 4.4 & 4.8  \\

        \textbf{Mistral-7B-Instruct-ZS} & 4.6 & 3.9 & 2.6 & 4.1  \\

        \textbf{Mistral-7B-Instruct-FT} & 4.8 & 4.0 & 2.4 & 4.0  \\

        \bottomrule
    \end{tabular}
     % %\vspace{-3mm}
    \caption{\small{Human evaluation results for Qualitative evaluation. Here, `ZS' and `FT' denote `Zero-Shot' and `Fine-Tuned', respectively.}}
    \label{tab:results_human_eval}
\end{table}

%\vspace{-1mm}

\paragraph{Preference Test:} We conduct a preference test by humans on 100 randomly sampled responses for the following 3 models that achieved 100\% format following accuracy in multi-query scenarios: GPT-4o, Gemini-1.5-Flash, and Claude-3.5-Sonnet. %The preference test is conducted by human reviewers on the responses generated in multi-query and single-query settings. 
Based on the preference test results demonstrated in Table \ref{tab:results_human_eval_preference}, we find that the summaries generated via multi-query prompting are preferred more by humans over the summaries generated via
single-query prompting for the evaluated LLMs.

\begin{table}[t!]
\scriptsize
\setlength{\tabcolsep}{4pt}
    \centering
    \begin{tabular}{l|ccc}
        \toprule
        \textbf{Model Name} & \textbf{Multi-Query Wins} & \textbf{Single-Query Wins} & \textbf{Tie} \\ % & %\textbf{FFC}  \\
        \midrule
        \textbf{GPT-4o}& 28.1\% & 3.1\% & 68.8\%  \\

        \textbf{Gemini-1.5-Flash} & 50.0\% & 6.3\% & 43.8\%  \\

        \textbf{Claude-3.5-Sonnet} & 56.3\% & 9.4\% & 34.3\%  \\    
\midrule
        \textbf{Average} & 44.8\% & 6.3\% & 48.9\%  \\  
        \bottomrule
    \end{tabular}
     % %\vspace{-3mm}
    \caption{\small{Human evaluation results for the Preference Test.}}
    \label{tab:results_human_eval_preference}
\end{table}

\subsection{Robustness}

In this section, we investigate the robustness of the proposed multi-query prompting approach in terms of the following:
(i) variations in instructions, (ii) different output formats, (iii)  generalizability on tasks beyond meeting summarization, and (iv) effects on optimized models. 
%\vspace{-1mm}
\paragraph{Instruction Variation:} Since there is a lack of query-focused meeting summarization datasets, we have used the QMSUM dataset for evaluation by converting it to the multi-query format. To investigate the robustness, we use the QMSUM-I dataset from \citet{fu-etal-2024-tiny} which is an instruction-focused version of the QMSUM dataset consisting of instructions to generate short/medium/long summaries. We consider the instructions for short/medium/long summary generation as individual queries and combine them together for the same transcript to construct a multi-query version. We find that all LLMs that achieve 100\% format following accuracy on QMSUM also achieve 100\% accuracy in QMSUM-I in the multi-query setting, with the best-performing open-source LLM, Mistral-7B-Instruct-v0.3, also maintaining a high format following accuracy of 95\%. However, LLMs that fail to achieve 100\% format-following accuracy in QMSUM also make errors in QMSUM-I. While the LLaMA-2-32K-Instruct again fails to generate any response in the proper format, we find that the performance in different datasets varies for other LLMs that achieve less than 100\% format following accuracy, as demonstrated in Figure \ref{fig:format_accuracy_qmsum_qmsumi}. We do not evaluate the results using automatic metrics like ROUGE or BERTScore 
in the QMSUM-I dataset since the reference summaries are synthetically generated using GPT-4. 

\begin{figure}[t!]
    \centering

        \includegraphics[width=\linewidth]{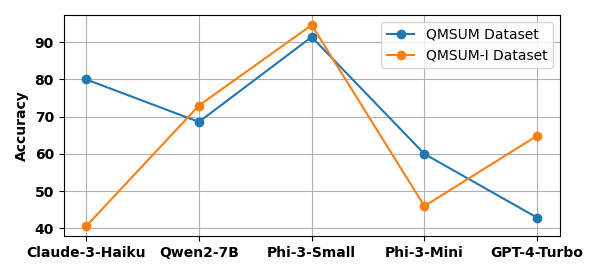}
  %\vspace{-3mm}
    \caption{\small{Format following capability of some zero-shot LLMs % that perform differently 
    in QMSUM and QMSUM-I datasets. LLMs achieving the same performance in both datasets are not shown here.}}

    \label{fig:format_accuracy_qmsum_qmsumi}
   
\end{figure}

\paragraph{Output Format Variation:}
%While we selected the best prompt for evaluation after comparing various prompts, 
We conduct further experiments to investigate the performance by prompting some LLMs (GPT-4o, Gemini-1.5-Flash, and Claude-3.5-Sonnet) that achieve 100\% format following accuracy in the ``JSON'' format to generate the response in the ``YAML'' format instead. Based on our experiments, we find that the format following accuracy is dropped to 97\%, 94\%, and 85\% from 100\% for GPT-4o, Gemini-1.5-Flash, and Claude-3.5-Sonnet, respectively. %Since the core focus of this work was to investigate the capabilities and limitations of various LLMs in the multi-query prompting setting for their real-world applicability in meeting summarization, 
This demonstrates that LLMs are more reliable in generating responses in ``JSON'' instead of ``YAML''. % since many LLMs demonstrated 100\% format following capability in this format. %Therefore, more extensive prompt engineering was left out of the scope of this research and suggested as a potential avenue for future research in the limitation section of our paper.

\paragraph{Out-of-domain Generalization:}
To investigate the out-of-domain generalization capability of different LLMs in our proposed multi-query prompting approach, we utilize the RACE-Hard \cite{lai-etal-2017-race} reading comprehension dataset in the multi-query setting and evaluate some of the LLMs studied in our paper. From the results stated in Table \ref{tab:results_generalizabilit}, we observe that multi-query prompting is successful in retaining high accuracy for many models, while models like LLaMA-2-7B-32k still struggle. % to output in the required format.

\begin{table}[t!]
\scriptsize
\setlength{\tabcolsep}{4pt}
    \centering
    \begin{tabular}{l|cc}
        \toprule
        \textbf{Model Name} & \textbf{JSON Format Following} & \textbf{QA Accuracy} \\ % & %\textbf{FFC}  \\
        \midrule
        \textbf{GPT-4o}& 100\% & 91.9 \\

        \textbf{Gemini-1.5-Flash} & 100\% & 89.0 \\

        \textbf{Claude-3-Haiku} & 99\% & 84.4  \\    

        \textbf{Mistral-7B-Instruct-V3} & 96\% & 70.8  \\ 
        \textbf{Phi-3-Mini-128K} & 62\% & 48.4 \\ 
        \textbf{LLaMA-2-7B-32K} & 2\% & 1.1   \\ 
        \bottomrule
    \end{tabular}
     % %\vspace{-3mm}
    \caption{\small{Generalizability of multi-query prompting beyond meeting summarization.}}
    \label{tab:results_generalizabilit}
\end{table}

\paragraph{Effects on Optimized Models:}
We apply 4-bit quantization in the best-performing open-source LLM: Mistral-7B-Instruct-v0.3 (in both zero-shot and fine-tuned) version and run inference using llama-cpp\footnote{\url{https://github.com/ggerganov/llama.cpp}} to investigate whether further optimization could still maintain the effectiveness in multi-query settings. Based on the results demonstrated in Table \ref{tab:results_optimization}, we observe that while the performance is degraded in zero-shot, the fine-tuned version could mostly retain the performance.

\begin{table}[t!]
\scriptsize
\setlength{\tabcolsep}{2pt}
    \centering
    \begin{tabular}{l|cccccc}
        \toprule
        \textbf{Model Name} & \textbf{Type} & \textbf{Format} & \textbf{R-1} & \textbf{R-2} 
        & \textbf{R-L} & \textbf{B-S}  \\ 
        & &  \textbf{Following} & & & & \\
        % & %\textbf{FFC}  \\
        \midrule

        \textbf{Mistral-7B-Instruct-ZS} & Original & 95\% & 30.1 & 9.4 & 20.3 & 59.8 \\

        \textbf{Mistral-7B-Instruct-ZS} & Quantized & 88\% & 27.2 & 9.1 & 19.0 & 57.7 \\
\midrule
        \textbf{Mistral-7B-Instruct-FT} & Original & 95\% & 30.0 & 10.1 & 20.9 & 59.9 \\

        \textbf{Mistral-7B-Instruct-FT} & Quantized & 95\% & 30.2 & 10.0 & 21.1 & 60.0  \\

        \bottomrule
    \end{tabular}
     % %\vspace{-3mm}
    \caption{\small{Effect on Optimized Models. Here, `ZS' and `FT' denote `Zero-Shot' and `Fine-Tuned', while `ROUGE' and `BERTScore' are denoted by `R' and `B-S', respectively.}}
    \label{tab:results_optimization}
\end{table}

%\vspace{-1mm}
\subsection{Usage Cost Analysis}
%\vspace{-1mm}
In this section, we demonstrate the benefit of multi-query prompting in terms of the usage cost. On average, each transcript in the QMSUM dataset has 8 corresponding queries. Thus, our proposed approach can reduce the cost 8X times in a dataset similar to QMSUM. For example, each transcript in QMSUM contains 9000 words on average, which is approximately 12000 tokens. Therefore, processing one transcript of 12000 tokens will cost\footnote{\url{https://openai.com/pricing}, last accessed: 18/07/2024.} 0.06 USD for GPT-4o. Thus, in the single-query scenario, if there are 8 queries for a single transcript, it would cost almost 0.48 USD. Thus, our approach could significantly save real-world usage costs. % while utilizing LLMs in real-world scenarios.

%\vspace{-1mm}
\section{Conclusion} 
%\vspace{-1mm}
In this paper, our experimental findings involving various LLMs led to several key insights on building an efficient real-world query-focused meeting summarization system. While most closed-source LLMs demonstrate superior performance in the multi-query setting, the open-source Mistral model with only 7B parameters performs on par with the closed-source LLMs, while also outperforming many of them. The relatively higher performance in terms of %BERTScore and ROUGE scores
automatic metrics in the multi-query setting, alongside significant optimization of cost and inference latency would open up the opportunities to utilize multi-query prompts in real-world industrial settings. 
%While we show that opting for multi-query prompting for LLM-based systems is a viable way to optimize the inference costs alongside retaining the performance in terms of automatic metrics.
In the future, we will study how to utilize tiny LLMs \cite{fu-etal-2024-tiny} similarly in the multi-query setting in the task of meeting summarization. %to generate the responses in the required format. % The multi-query version of the QMSUM dataset and the code used in this research would be made publicly available.

% Future work should investigate how to make more LLMs to generate the responses in the required format to ensure better parsing of the LLM-generated outputs. 

\section*{Acknowledgements}
We would like to thank the program chairs, area chairs, and the anonymous reviewers for reviewing the paper and providing valuable feedback. We also thank our colleagues \textbf{Gundeep Singh} and \textbf{Julien Bouvier Tremblay} for their help with the human evaluation study. 

\section*{Limitations}
One of the limitations of this work is that in the multi-query setting, since many LLMs could not produce the outputs in the required format, we only use those summaries for evaluation that could be extracted using our custom parser. While we carefully designed the parser such that it could handle all possible response types to extract the summary, it was not possible in some cases to extract the corresponding summary for a given query due to the variance in LLM-generated responses. We did not enlist human help to extract the summary in these cases since our goal is to build this multi-query prompting for real-world industrial scenarios that require automatic parsing of the corresponding summary for a given input. Nonetheless, future work should focus on improving the instruction-following capability of LLMs in terms of the output format. 

Further, more extensive prompt engineering could be beneficial. While we selected the prompt for evaluation after comparing various prompts, it still did not help these LLMs to generate properly formatted output. Nonetheless, our experiments were limited to only zero-shot prompting and fine-tuning. % to generate responses in JSON format. 
Therefore, %other types of formatting, such as YAML \cite{xia2024fofo}, alongside 
investigating advanced prompting techniques \cite{schulhoff2024prompt}, such as Chain-of-Thought (CoT) \cite{wei2022chainofthough}, few-shot learning, etc. may help improve the response generation quality of many LLMs that should be studied in the future. 

%Moreover, this work only uses the original models to evaluate their ability to handle multi-purpose queries. Applying optimization techniques \cite{zhu2023survey} like post-training quantization, quantization-aware training, etc. are left out of scope for this paper, but could be a good direction for future work. 

\section*{Ethics Statement}

%\paragraph{License:} 
We maintained the licensing requirements accordingly while using different tools from the providers (e.g., OpenAI, Google, Meta, Microsoft, HuggingFace). Only a publicly available academic dataset has been used which is available for both research and commercial purposes. All the human evaluations were done by our in-house computational linguists and so additional compensations were not required. 

%\paragraph{Human Evaluation:} Additional compensations were not required for the human evaluation since it was conducted by in-house full-time employees having expertise in computational linguistics.

\bibliography{anthology,custom}
\bibliographystyle{acl_natbib}

\appendix

\section{Appendix}
\label{appendix}

In this section, we show some examples of error responses in different LLMs. 

\begin{itemize}
    \item See Table \ref{table:appendix_claude_3_haiku} for Claude-3-Haiku.
     \item See Table \ref{table:appendix_qwen2} for Qwen2.
      \item See Table \ref{table:appendix_phi_3_small} for Phi-3-Small.
       \item See Table \ref{table:appendix_phi_3_mini} for Phi-3-Mini.
        \item See Table \ref{table:appendix_gpt_4_turbo} for GPT-4-Turbo.
\end{itemize}

 \begin{table*}[t!]
\centering
\scriptsize
\begin{tabular}{p{7cm}|p{6cm}|p{2cm}}
\toprule %0.7009676372					0.6792449793						0.7009676372	0.6899353658
  %  \textbf{Model} & 
 \textbf{Prompt} & \textbf{Claude-3-Haiku Response} & \textbf{Comments} \\ 
\midrule
%GPT-3.5 ($n$ = 5000) &         truncation (2 Sentence) & 31.48 & 13.32 & 23.64 & 60.25 &         49.60\\
A list of queries followed by a transcript is given below. For each of the following queries, generate the query-focused summary of the given transcript in an Array of JSON objects. You must give your response only in the required Array of JSON objects format and your response for each JSON object should contain the corresponding values for the following keys: (i) query and (ii) summary.  \newline

\#Queries Begin  \newline
1. Summarize the whole meeting.  \newline
2. What did the group discuss the functional design on the remote control?  \newline
3. What did the group discuss the finder button and call button?  \newline
4. What did group members think of the speech recognition design suggested by Marketing?  \newline
5. Summarize the discussion about the VCR / VHR / DVD capability button.  \newline
\#Queries End  \newline

\#Queries End  \newline

\#Transcript Begin  \newline
[Transcript]  \newline
\#Transcript End  \newline

& \textcolor{red}{\#1}  \newline
{  \newline
  "query": "Summarize the whole meeting.",  \newline
  "summary": "The group discussed the functional design of a remote control for a TV. They reviewed user research findings on remote control usage habits, and decided to focus on the most frequently used buttons like channel selection, volume, and power. They debated incorporating features like speech recognition and a locator function, but decided those may be too complex or expensive. The group discussed the remote's button layout and design, considering a touch-based scrolling interface similar to an iPod. They also discussed whether to include DVD and VCR functionality, and decided to focus primarily on TV controls. Overall, the group aimed to create a user-friendly, fashionable remote control targeted at teenagers and young professionals."  \newline
} \newline 

\textcolor{red}{\#2} \newline
{ \newline
  "query": "What did the group discuss the functional design on the remote control?",  \newline
  "summary": "The group discussed the key functional requirements for the remote control, including being able to turn the TV on/off, change channels, and adjust volume. They also considered adding a locator function to help find the remote if misplaced, which would involve a separate base station that could send a signal to make the remote beep or vibrate. The group debated incorporating more advanced features like speech recognition, but decided those may be too complex or expensive to implement."  \newline
} \newline

\textcolor{red}{\#3} \newline
{ \newline
  "query": "What did the group discuss the finder button and call button?", \newline
  "summary": "The group discussed implementing a locator function where there would be a separate base station that could send a signal to make the remote control beep or vibrate, to help find it if misplaced. They decided this would be a separate component sold with the remote, rather than integrated into the remote itself, as otherwise you would need to find the remote first in order to locate it. The group agreed that a simple beep sound would be the best output method, rather than a light, as the sound could be heard even if the remote was hidden." \newline
} \newline

\textcolor{red}{\#4} \newline
{ \newline
  "query": "What did group members think of the speech recognition design suggested by Marketing?", \newline
  "summary": "The group was skeptical about implementing speech recognition functionality on the remote control. They noted that it would be technically challenging to program, as the TV audio could interfere with voice commands. Additionally, they were concerned about the cost of adding speech recognition to every remote, given their target budget. Overall, the group decided not to pursue speech recognition, focusing instead on more basic and cost-effective features." \newline
} \newline

\textcolor{red}{\#5} \newline
{ \newline
  "query": "Summarize the discussion about the VCR / VHR / DVD capability button.",
  "summary": "The group discussed whether to include VCR and DVD functionality on the remote control, in addition to TV controls. They noted that VCR functionality would be challenging, as VCR standards differ internationally. However, they felt that including DVD controls could be useful, as DVD players are commonly used with TVs. Ultimately, they decided to focus the remote primarily on TV controls, as most users would likely have separate remotes for DVD players. The group considered having a mode switch button to toggle between TV, DVD, and potentially other device controls, but did not reach a final decision on this."
} \newline
] & {Inappropriate \newline format.} \\
\bottomrule
\end{tabular}
\caption{\small{Claude-3-Haiku generated response.}}
\label{table:appendix_claude_3_haiku}
\end{table*}

 \begin{table*}[t!]
\centering
\scriptsize
\begin{tabular}{p{7cm}|p{6cm}|p{2cm}}
\toprule %0.7009676372					0.6792449793						0.7009676372	0.6899353658
  %  \textbf{Model} & 
 \textbf{Prompt} & \textbf{Qwen2 Response} & \textbf{Comments} \\ 
\midrule
%GPT-3.5 ($n$ = 5000) &         truncation (2 Sentence) & 31.48 & 13.32 & 23.64 & 60.25 &         49.60\\
A list of queries followed by a transcript is given below. For each of the following queries, generate the query-focused summary of the given transcript in an Array of JSON objects. You must give your response only in the required Array of JSON objects format and your response for each JSON object should contain the corresponding values for the following keys: (i) query and (ii) summary.

\#Queries Begin \newline
1. What was the overall discussion of the meeting?  \newline
2. What were some of the ideas proposed about future meeting  recordings?  \newline
3. What was said on getting fluent English speakers?  \newline
4. What were the options that were discussed on the location of the recording equipment?  \newline
5. What did the group say on portable equipment?  \newline
6. What were Grad B’s updates on meeting disk storage? \newline
7. What were other ways to get more space?   \newline
\#Queries End

\#Transcript Begin  \newline
[Transcript]  \newline
\#Transcript End  \newline

 &   
What is PhD C's main research goal regarding the overlap issue?
& {Possibly a hallucinated answer that is not at all relevant to the context.} \\
\bottomrule
\end{tabular}
\caption{\small{Qwen2 generated response.}}
\label{table:appendix_qwen2}
\end{table*}

 \begin{table*}[t!]
\centering
\scriptsize
\begin{tabular}{p{7cm}|p{6cm}|p{2cm}}
\toprule %0.7009676372					0.6792449793						0.7009676372	0.6899353658
  %  \textbf{Model} & 
 \textbf{Prompt} & \textbf{Phi-3-Small Response} & \textbf{Comments} \\ 
\midrule
%GPT-3.5 ($n$ = 5000) &         truncation (2 Sentence) & 31.48 & 13.32 & 23.64 & 60.25 &         49.60\\
A list of queries followed by a transcript is given below. For each of the following queries, generate the query-focused summary of the given transcript in an Array of JSON objects. You must give your response only in the required Array of JSON objects format and your response for each JSON object should contain the corresponding values for the following keys: (i) query and (ii) summary.  \newline

\#Queries Begin  \newline
1. Summarize the whole meeting.  \newline
2. What did the group discuss the functional design on the remote control?  \newline
3. What did the group discuss the finder button and call button?  \newline
4. What did group members think of the speech recognition design suggested by Marketing?  \newline
5. Summarize the discussion about the VCR / VHR / DVD capability button.  \newline
\#Queries End  \newline

\#Queries End  \newline

\#Transcript Begin  \newline
[Transcript]  \newline
\#Transcript End  \newline

&  Based on the provided transcript, here are the JSON objects summarizing the key points of the meeting: \newline

[ \newline
  \{ \newline
    "query": "Summarize the whole meeting.",
    "summary": "The meeting focused on the functional design of a remote control, discussing user needs, technical effects, and the working design. The team reviewed user testing results, button usage statistics, and preferences for remote control features. Marketing presented findings on user preferences and willingness to pay for features like speech recognition. The team discussed the importance of focusing on frequently used buttons and the potential for a simplified remote design. Technical functions and design considerations for the remote were also presented, including the possibility of incorporating a locator button and speech recognition. The team debated the inclusion of VCR/VHR/DVD capabilities and the need for a locator function. The meeting concluded with a discussion on the target demographic and the importance of a fashionable design."
  \}, \newline
  \{ \newline
    "query": "What did the group discuss the functional design on the remote control?",
 & {Incomplete response leading to inappropriate formatting.} \\
\bottomrule
\end{tabular}
\caption{\small{Phi-3-Small generated response.}}
\label{table:appendix_phi_3_small}
\end{table*}

 \begin{table*}[t!]
\centering
\tiny
\begin{tabular}{p{7cm}|p{6cm}|p{2cm}}
\toprule %0.7009676372					0.6792449793						0.7009676372	0.6899353658
  %  \textbf{Model} & 
 \textbf{Prompt} & \textbf{Phi-3-Mini Response} & \textbf{Comments} \\ 
\midrule
%GPT-3.5 ($n$ = 5000) &         truncation (2 Sentence) & 31.48 & 13.32 & 23.64 & 60.25 &         49.60\\
A list of queries followed by a transcript is given below. For each of the following queries, generate the query-focused summary of the given transcript in an Array of JSON objects. You must give your response only in the required Array of JSON objects format and your response for each JSON object should contain the corresponding values for the following keys: (i) query and (ii) summary.
\newline \#Queries Begin \newline 
1. Summarize the meeting \newline
2. What’s the thesis about? \newline
3. What was said about EML talk? \newline
4. What was said about the talk and neurons?
\newline \#Queries End \newline 

\#Transcript Begin \newline
[Transcript] \newline
\#Transcript End \newline
&   [ \newline
  \{"query": "Summarize the meeting", "summary": "The meeting started with a casual greeting and a brief discussion about the comfort of the room. Professor E mentioned that Keith and the speaker were a bit dazed due to being recorded. The group then moved on to discuss the speaker's thesis proposal, with Professor E suggesting that the proposal could be improved by emphasizing the scalability of combining constructions with a general construal mechanism. The group also discussed the importance of metonymy and metaphor in the tourism domain, with Professor E providing examples of how these linguistic concepts could be applied. The meeting concluded with Professor E encouraging the speaker to seek feedback electronically and wishing them a successful trip."\}
\textcolor{red}{]} \newline
\textcolor{red}{[} \newline
  {"query": "What’s the thesis about?", "summary": "The thesis is about developing a probabilistic relational model that uses ontology, situation, user, and discourse models to make inferences about constructions in the tourism domain. The speaker also mentioned working on a tablet-based parser and a synthesis system, and plans to discuss data, problems with rephrasing, preliminary observations, and the Bayes-net in their talk."}
\textcolor{red}{]} \newline
\textcolor{red}{[} \newline
  {"query": "What was said about EML talk?", "summary": "The speaker plans to discuss their work on SmartKom, including a parser and a generation outputter, and their research on language understanding. They will also talk about their problems with rephrasing, preliminary observations, and the NTL story. The speaker will also discuss the embodied and simulation approach, monkeys in Italy, and future activities related to FMRI studies and mirror neurons. The speaker will also present a talk at EML, focusing on ontological knowledge in the tourism domain and the use of X-schemas. The speaker will also discuss the importance of metonymy and metaphor in the domain and plans to seek feedback electronically."}
\textcolor{red}{]} \newline
\textcolor{red}{[/JSONObjects]} \newline

] & Inappropriate formatting. Wrongly additional brackets and texts. \\
\bottomrule
\end{tabular}
\caption{\small{Phi-3-Mini generated response.}}
\label{table:appendix_phi_3_mini}
\end{table*}

 \begin{table*}[t!]
\centering
\tiny
\begin{tabular}{p{7cm}|p{6cm}|p{2cm}}
\toprule %0.7009676372					0.6792449793						0.7009676372	0.6899353658
  %  \textbf{Model} & 
 \textbf{Prompt} & \textbf{GPT-4-Turbo Response} & \textbf{Comments} \\ 
\midrule
%GPT-3.5 ($n$ = 5000) &         truncation (2 Sentence) & 31.48 & 13.32 & 23.64 & 60.25 &         49.60\\
A list of queries followed by a transcript is given below. For each of the following queries, generate the query-focused summary of the given transcript in an Array of JSON objects. You must give your response only in the required Array of JSON objects format and your response for each JSON object should contain the corresponding values for the following keys: (i) query and (ii) summary.
\newline \#Queries Begin \newline
1. Summarize the meeting
2. What’s the thesis about?
3. What was said about EML talk?
4. What was said about the talk and neurons?
\newline \#Queries End

\#Transcript Begin
[Transcript]
\#Transcript End
&   [ \newline
    \{ \newline
        "query": "Summarize the meeting",
        "summary": "The meeting involved discussions about a thesis proposal, rehearsing a talk for EML, and suggestions for improving an upcoming presentation. Grad C discussed incorporating feedback for his thesis on construal and language metaphors, needing to update it to include a general construal mechanism. They also discussed potentially using construal to handle metaphorical language in computational systems. There were tangents regarding office designs at EML and how people speak to digital systems, but these served as background to substantiate research discussions."
    \}, \newline
    \{ \newline
        "query": "What’s the thesis about?",
        "summary": "The thesis discussed in the meeting revolves around applying a general construal mechanism to interpret metaphorical language in computational systems. It proposes that by adding this mechanism, computational systems can handle not only basic language structures but also metaphorical expressions more efficiently. The thesis is intended to demonstrate how computational systems could expand their understanding through modular constructions in grammar that interact with metaphorical and metonymic uses of language."
    \}, \newline
   \{ \newline
 \textcolor{red}{.getText":} "What was said about EML talk?",
        "summary": "Grad C plans to give a talk at EML about the project's achievements, which include early work on the SmartKom system and their insights from processing human-computer interactions. He intends to use practical examples of how ontological knowledge and context influence interpretation of queries in computational systems, using hypothetical tourist scenarios and current real-life applications. Additionally, he discussed pulling in various components for the talk, including slides on neurology and the construction simulation model."
    \}, \newline
    \{ \newline
        \textcolor{red}{"text":} "What was said about the talk and neurons?",
        "summary": "In the discussion about the upcoming talk, the Professor suggested mentioning the newer research on mirror neurons when talking about embodied and simulation approaches to language understanding. This inclusion is intended to align with their broader research on how humans process language and could enhance the overall presentation by linking neural activities to computational models of language understanding."
    \} \newline
] & Wrong JSON Keys (should have been `query'). \\
\bottomrule
\end{tabular}
\caption{\small{GPT-4-Turbo generated response.}}
\label{table:appendix_gpt_4_turbo}
\end{table*}

\end{document}